\documentclass[lettersize,journal]{IEEEtran}
\usepackage{amsmath,amsfonts}
\usepackage{algorithmic}
\usepackage{algorithm}
\usepackage{array}
\usepackage[caption=false,font=normalsize,labelfont=sf,textfont=sf]{subfig}
\usepackage{textcomp}
\usepackage{stfloats}
\usepackage{url}
\usepackage{verbatim}
\usepackage{graphicx}
\usepackage{cite}
\usepackage{array}
\usepackage[caption=false,font=normalsize,labelfont=sf,textfont=sf]{subfig}
\usepackage{textcomp}
\usepackage{stfloats}
\usepackage{url}
\usepackage{verbatim}
\usepackage{graphicx}
\usepackage{cite}
\usepackage{tikz,xcolor}
\usepackage{orcidlink}
\usepackage{times}
\usepackage{epsfig}
\usepackage{graphicx}
\usepackage{amsmath}
\usepackage{amssymb}
\usepackage{booktabs}
\usepackage{algorithm}
\usepackage{algorithmic}
\usepackage{makecell}
\usepackage{cuted}
\usepackage{capt-of}
\usepackage{multirow}
\usepackage{caption}
\hypersetup{hidelinks,
colorlinks=true,
allcolors=black,
pdfstartview=Fit,
breaklinks=true
}
\usepackage{lettrine}
\usepackage{lipsum}
\usepackage{enumitem}

\begin{document}

\title{Isolated Diffusion: Optimizing Multi-Concept Text-to-Image Generation Training-Freely with Isolated Diffusion Guidance}

\author{Jingyuan Zhu, Huimin Ma, Jiansheng Chen, and Jian Yuan}



\maketitle

\begin{abstract}
Large-scale text-to-image diffusion models have achieved great success in synthesizing high-quality and diverse images given target text prompts. Despite the revolutionary image generation ability, current state-of-the-art models still struggle to deal with multi-concept generation accurately in many cases. This phenomenon is known as ``concept bleeding" and displays as the unexpected overlapping or merging of various concepts. This paper presents a general approach for text-to-image diffusion models to address the mutual interference between different subjects and their attachments in complex scenes, pursuing better text-image consistency. The core idea is to isolate the synthesizing processes of different concepts. We propose to bind each attachment to corresponding subjects separately with split text prompts. Besides, we introduce a revision method to fix the concept bleeding problem in multi-subject synthesis. We first depend on pre-trained object detection and segmentation models to obtain the layouts of subjects. Then we isolate and resynthesize each subject individually with corresponding text prompts to avoid mutual interference. Overall, we achieve a training-free strategy, named Isolated Diffusion, to optimize multi-concept text-to-image synthesis. It is compatible with the latest Stable Diffusion XL (SDXL) and prior Stable Diffusion (SD) models. We compare our approach with alternative methods using a variety of multi-concept text prompts and demonstrate its effectiveness with clear advantages in text-image consistency and user study.
\end{abstract}


\section{Introduction}
\label{sec:intro}

In recent years, diffusion probabilistic models \cite{sohl2015deep,NEURIPS2020_4c5bcfec,song2020improved, dhariwal2021diffusion, nichol2021improved,kingma2021variational} have attracted significant attention from both academia and industry with their outstanding performance and application to various downstream tasks \cite{ruiz2023dreambooth,gal2022itextual,kumari2023multi,gu2023mix,raj2023dreambooth3d,zhu2023domainstudio,kawar2023imagic}. Modern large-scale text-to-image diffusion models including Imagen \cite{saharia2022photorealistic}, DALL·E \cite{ramesh2022hierarchical}, eDiff-I \cite{balaji2022ediffi}, and Stable Diffusion \cite{rombach2022high,podell2023sdxl} optimize numerous parameters using billions of training data and demonstrate unparalleled capabilities to produce high-quality and diverse samples given unfettered text prompts.

\begin{figure}[t]
    \centering
    \includegraphics[width=1.0\linewidth]{ 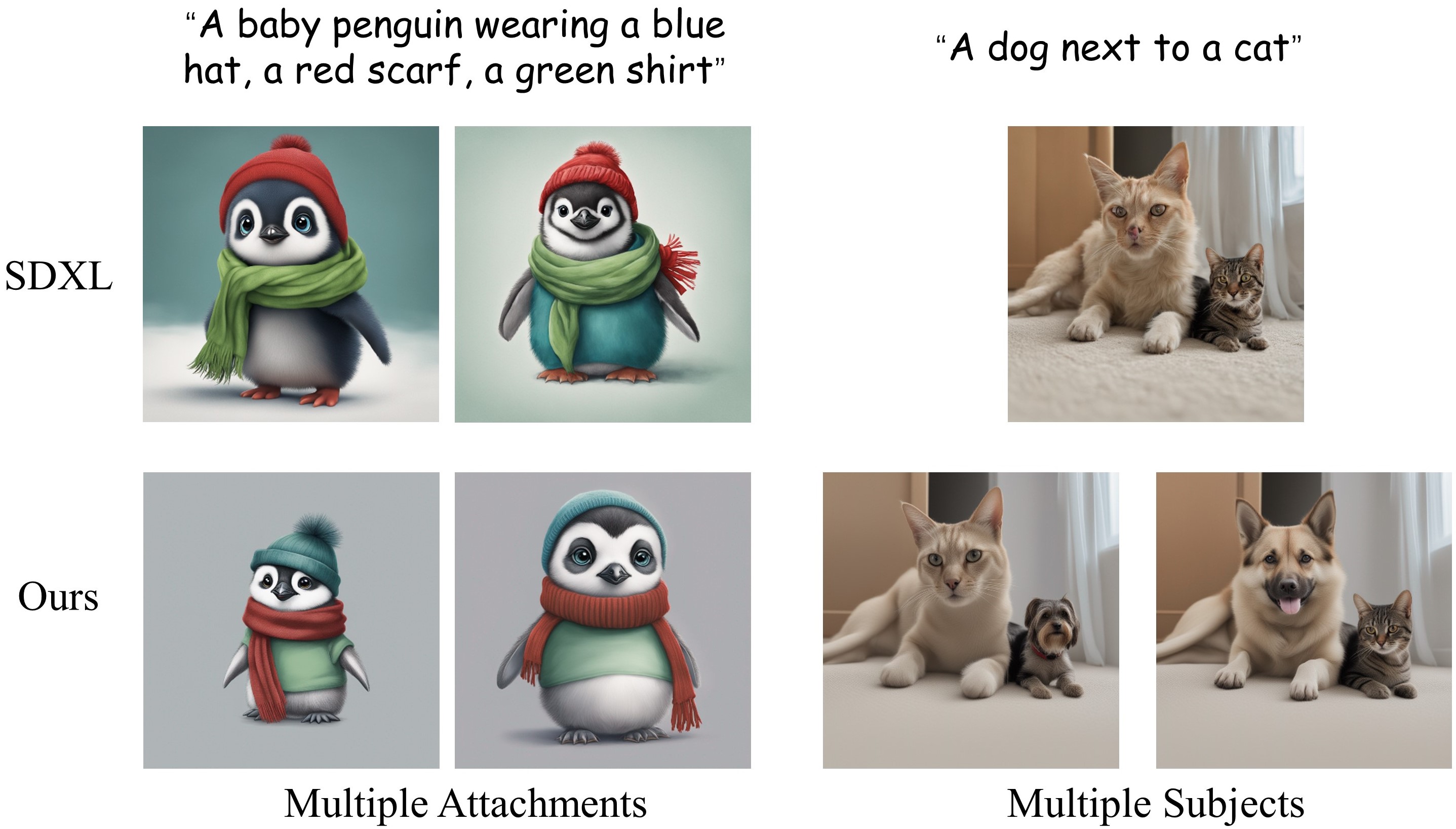}
    \caption{Multi-concept generated samples comparison. Our approach fixes the concept bleeding problem of SDXL.}
    \label{examples0}
\end{figure}

Despite the compelling results of modern text-to-image models in single-concept cases, they still suffer from text-image inconsistency given complex text prompts containing multiple concepts. Taking the latest SD model SDXL as an example, it has made great progress in avoiding the concept missing problem compared with prior SD models. However, SDXL still exhibits a phenomenon named ``concept bleeding" to synthesize images inconsistent with text prompts in some multi-concept synthesis cases. Concept bleeding is considered to be caused by the pre-trained text encoders \cite{radford2021learning,bigclip}, which compress all information in the text prompt into a specific number of tokens. As a result, different concepts in text prompts may interfere with each other in the encoding process. As shown in Fig. \ref{examples0}, SDXL assigns the colors of various attachments erroneously in the left samples and merges the concept of ``a cat" to both subjects in the right sample, resulting in unreasonable results.

A series of methods have been proposed to improve multi-concept generation. Composable Diffusion \cite{liu2022compositional} first attempts to solve this problem by synthesizing different components in an image with a set of diffusion models using text prompts split according to conjunctions. It struggles to handle the interference between various concepts since it adds up noises predicted with split prompts directly. Following works \cite{liu2022compositional,feng2022training,chefer2023attend,mao2023training,chen2023training,ma2023directed,wu2023harnessing,li2023divide,rassin2024linguistic,meral2023conform} optimize cross-attention maps or latents to enhance text-image consistency. Some of them \cite{mao2023training,chen2023training,ma2023directed,wu2023harnessing} introduce additional controls like layouts to enhance each concept. However, they still utilize embeddings encoded from complete text prompts, leading to inevitable concept bleeding in many cases.

\begin{figure}[t]
    \centering
    \includegraphics[width=1.0\linewidth]{ 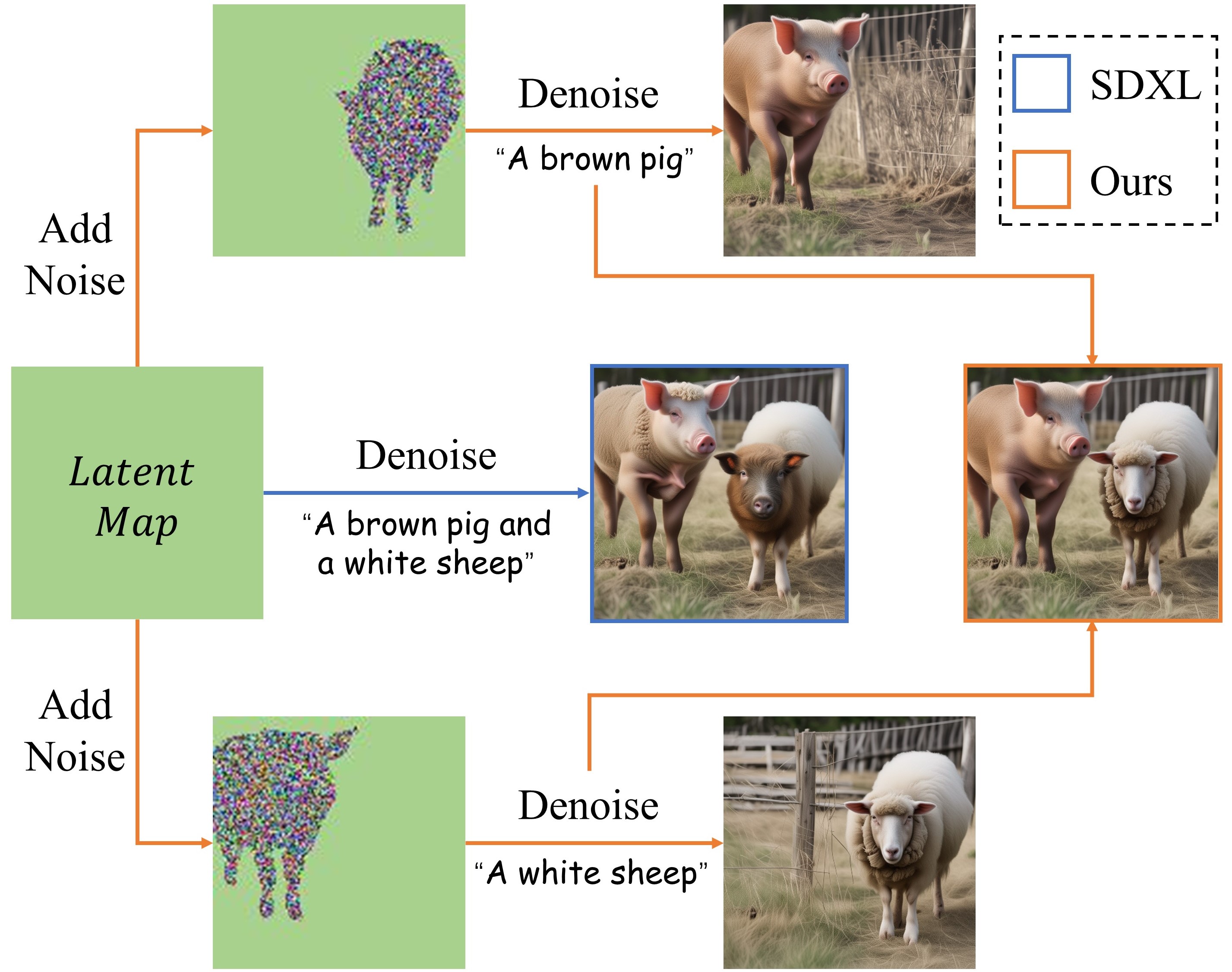}
    \caption{We isolate multi-subject generation by replacing the regions of other subjects in latents with random noises and denoise each subject with split text prompts individually. Our approach avoids mutual interference between various subjects and gets a more reasonable result than SDXL.}
    \label{pipeline0}
\end{figure}

In this paper, we propose training-free approaches based on the open-source SD models \cite{rombach2022high,podell2023sdxl} to deal with two typical challenges in multi-concept generation: concept bleeding of multiple attachments and subjects, as shown in Fig. \ref{examples0}. The key idea is to isolate the denoising processes of various concepts to relieve mutual interference. For multiple attachments, we use isolated text prompts to bind each attachment to corresponding subjects separately. For multiple subjects, we first segment the concept bleeding samples with pre-trained detection and segmentation models like YOLO \cite{wang2023yolov7} and SAM \cite{kirillov2023segment} to identify each subject and their layouts with masks. Then we figure out an effective method to resynthesize each subject individually using corresponding text prompts by replacing the regions of other subjects with random noises. Taking Fig. \ref{pipeline0} as an example, we replace the region of ``pig" in the latent with random noises and denoise ``sheep" with the text prompt ``A white sheep". In this way, we revise the ``sheep" with a brown pig head produced by SDXL to a white sheep consistent with the text prompt. We denote our approach as Isolated Diffusion. The main contributions are concluded as follows:
\begin{itemize}[leftmargin=0.4cm]
    \item We propose an intuitive inference method to bind each attachment to corresponding subjects separately and improve the text-image consistency of multi-attachment synthesis.
    \item We design a training-free method to revise multi-subject samples with pre-trained detection and segmentation models (e.g., YOLO and SAM) to keep image layouts and avoid unexpected mutual interference between various subjects, achieving better text-image consistency.
    \item We conduct sufficient experiments and user study to demonstrate the effectiveness of our approach.
\end{itemize}

\section{Related Work}
\label{sec:related}
\textbf{Diffusion Models} Diffusion models \cite{sohl2015deep,NEURIPS2020_4c5bcfec,song2020improved, dhariwal2021diffusion, nichol2021improved,kingma2021variational} have become mainstream in image synthesis, outperforming GANs \cite{NIPS2014_5ca3e9b1, DBLP:conf/iclr/BrockDS19, Karras_2019_CVPR, Karras_2020_CVPR, Karras2021}, VAEs \cite{kingma2013auto, rezende2014stochastic, vahdat2020nvae}, and autoregressive models \cite{van2016conditional,chen2018pixelsnail,henighan2020scaling} on both generation quality and diversity. Moreover, a series of works \cite{zhang2023adding,zhao2023uni,mou2023t2i,li2023gligen,nichol2022glide,wang2022semantic,epstein2023diffusion} have introduced a variety of controls to realize diffusion-based conditional generation, such as text, sketch \cite{simo2016learning,simo2018mastering}, edge \cite{canny1986computational,gu2022towards}, and segmentation mask \cite{xiao2018unified} et al. As a result, diffusion models are applicable to all kinds of tasks, including text-to-image synthesis \cite{rombach2022high,podell2023sdxl,saharia2022photorealistic,ramesh2022hierarchical,balaji2022ediffi,gafni2022make}, layout-to-image generation \cite{cheng2023layoutdiffuse,zheng2023layoutdiffusion,xue2023freestyle,li2023generate,xie2023boxdiff,bar2023multidiffusion,zhang2023adding,li2023gligen}, inpainting \cite{xie2023smartbrush,lugmayr2022repaint}, outpainting \cite{wu2023ipo}, super resolution \cite{saharia2022image}, and customization of subjects and styles \cite{ruiz2023dreambooth,gal2022itextual,kumari2023multi,zhu2023domainstudio,gu2023mix,ruiz2023hyperdreambooth,raj2023dreambooth3d,ma2023subject,ahn2023dreamstyler}.

\textbf{Text-to-Image Synthesis}
Diffusion-based text-to-image synthesis attracts the most visits among all the conditional generation tasks, benefiting from universal natural languages and several widely-used large-scale generative models \cite{saharia2022photorealistic,ramesh2022hierarchical,balaji2022ediffi,rombach2022high}. Our work is implemented with SDXL \cite{podell2023sdxl}, the latest version of SD models \cite{rombach2022high}. SD is a two-stage text-to-image generation model composed of pre-trained autoencoders and a transformer-based UNet, which works on low-resolution latents encoded by the autoencoders. Compared with prior open-source SD models like SD1.5 and SD2.1, SDXL improves the scale of parameters in its UNet from about 860M to 2.6B and introduces an optional refiner network to improve generation quality. In addition, SDXL employs two powerful text encoders, OpenCLIP ViT-bigG \cite{bigclip} and CLIP ViT-L \cite{radford2021learning}, to encode text prompts. Nevertheless, SDXL still cannot completely avoid concept bleeding in multi-concept generation.

\textbf{Concept Bleeding}
Apart from the pioneer Composable Diffusion \cite{liu2022compositional}, other works focus on manipulating cross-attention maps or latents. Structured Diffusion \cite{feng2022training} encodes word relationships into the encoding explicitly. Attend-and-Excite \cite{chefer2023attend} slightly shifts the latents and refines the cross-attention units with all subject tokens to avoid subjects missing. Divide-and-Bind \cite{li2023divide} optimizes latents based on losses computed with cross-attention maps. SynGen \cite{rassin2024linguistic} guides cross-attention maps to align with syntactically analyzed prompts. Other methods \cite{mao2023training,chen2023training,ma2023directed,wu2023harnessing,meral2023conform} manipulate cross-attention maps using the prior knowledge of each concept's layout with or without additional training or networks. Our work introduces an intuitive and training-free approach to isolate various concepts without directly manipulating attention maps.

\textbf{Image Editing} Diffusion-based text-driven image editing \cite{bau2021paint,nichol2022glide,avrahami2023blended,huang2023region,mao2023guided,brooks2023instructpix2pix,couairon2022diffedit,avrahami2022blended} methods tackle a task different from this work. They manipulate part of the given images and keep the other parts unchanged. We handle the concept bleeding problem of SD models to achieve text-image consistency regardless of the consistency between revised and original samples. It is hard to compare our work with them directly. Our approach for multiple subjects synthesizes subjects within ranges of their masks separately. Blended Diffusion \cite{avrahami2023blended,avrahami2022blended} denoise the whole image with text prompts of edited parts and merge foreground and background with masks, leading to unrealistic results. DiffEdit \cite{couairon2022diffedit} provides more accurate mask guidance but still struggles to deal with multi-subject scenes. We denoise background with latents containing foreground information and achieve realistic results with high text-image consistency.

\section{Method}
\label{sec:approach}
In this section, we present the Isolated Diffusion approach to relieve two typical concept bleeding problems of SD models in multi-concept generation. Isolated Diffusion for multiple attachments and multiple subjects are introduced in detail separately in Sec. \ref{32} and \ref{33}.

\subsection{Isolated Diffusion for Multiple Attachments}
\label{32}
\begin{figure*}[p]
    \centering
    \includegraphics[width=1.0\linewidth]{ 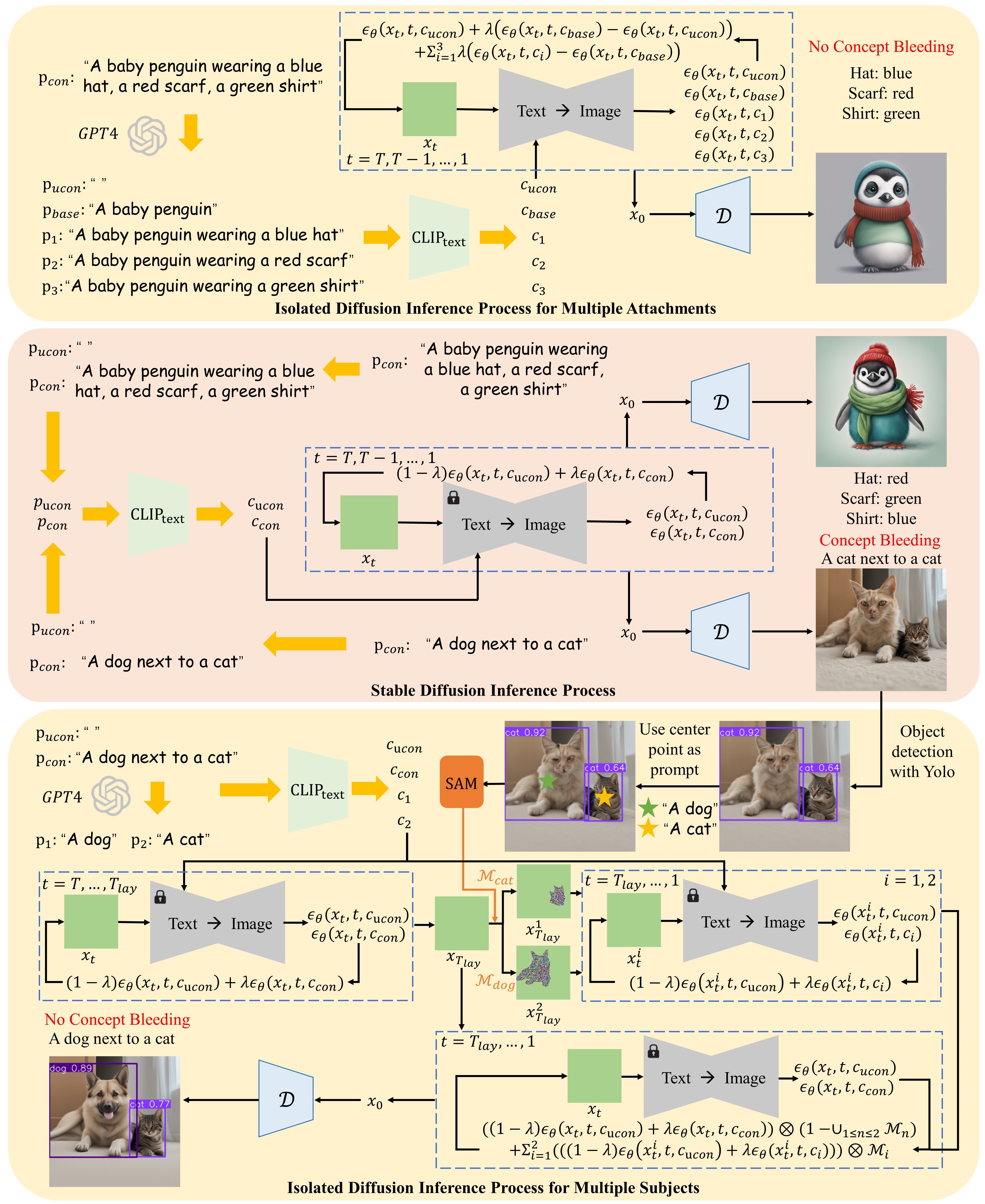}
    \caption{Overview of Isolated Diffusion. We decompose complex text prompts into simpler forms with GPT4 and denoise each concept under split conditions to avoid mutual interference between various concepts for better text-image consistency.}
    \label{pipeline}
\end{figure*}

Given text prompts containing multiple attachments bound to a single subject, the current SD inference strategy struggles to maintain considerable text-image consistency. For example, a series of attachments and corresponding color descriptions exist in the text prompt ``A baby penguin wearing a blue hat, a red scarf, and a green shirt". As shown in the middle part of Fig. \ref{pipeline}, SD encounters concept bleeding and produces samples inconsistent with the text prompt. It is considered to be caused by the pre-trained CLIP models \cite{bigclip,radford2021learning}, which encode complex text prompts altogether into a specific number of tokens as the condition $c_{con}$. Therefore, the color descriptions are mixed up among multiple attachments. The predicted noises $\hat{\epsilon}(x_t,t)$ in the current inference process are defined as the linear combination of the unconditionally predicted noises $\epsilon_{\theta}(x_t,t,c_{ucon})$ and conditionally predicted noises $\epsilon_{\theta}(x_t,t,c_{con})$:
\begin{align}
    \hat{\epsilon}(x_t,t) = (1-\lambda)\epsilon(x_t,t,c_{ucon}) + \lambda \epsilon(x_t,t,c_{con}),
\end{align}
where $\lambda$ represents the scaling on the condition $c_{con}$.

To overcome the concept bleeding problem of multiple attachments, Isolated Diffusion first splits complex text prompts into a group of simpler text prompts with GPT4 \cite{schick2023toolformer}, including a base subject $\text{p}_{base}$ and the base subject bound with each attachment $\text{p}_1, \text{p}_2, \ldots, \text{p}_k$ separately. Isolated Diffusion conducts the denoising process using the linear combination of the noises predicted under the split conditions. We first add the variance between the noises predicted unconditionally and with $\text{p}_{base}$ to synthesize the base subject. Then, we add the variance between the noises predicted under the prompt of the subject and the prompt of a single attachment bound to the subject to provide separate guidance. The overall predicted noises of Isolated Diffusion can be expressed as follows:
\begin{equation}
    \begin{aligned}
    \hat{\epsilon}(x_t,t)&= (1-\lambda) \epsilon_{\theta}(x_t,t,c_{ucon}) + \lambda \epsilon_{\theta}(x_t,t,c_{base})  \\
    & + \sum_{i=1}^k \lambda(\epsilon_{\theta}(x_t,t,c_{i})-\epsilon_{\theta}(x_t,t,c_{base})) \label{eq2}
\end{aligned}
\end{equation}

In this way, we avoid interference between multiple attachments. The pseudo-code of Isolated Diffusion for multiple attachments is provided in Algorithm \ref{algo:algo1}. Compared with the current SD inference process, we only need to split the complex text prompts without additional training. As illustrated in the top part of Fig. \ref{pipeline}, our approach achieves accurate assignments of color descriptions to multiple attachments consistent with the text prompt without fidelity degradation.

Our approach shares a similar idea of splitting the synthesizing processes of different components in an image with Composable Diffusion \cite{liu2022compositional}. The core difference is that our approach binds each attachment to the base subject individually, while Composable Diffusion adds up the predicted noises of subjects and attachments directly. Given a text prompt ``A table with red table cloth and yellow tulips", Composable Diffusion decomposes it into a group of text prompts: ``a table", ``red table cloth", and ``yellow tulips" and adds up the noises predicted with these split text prompts directly. However, these noises may overlap or influence each other in the denoising process, leading to concepts missing or merging. We decompose the text prompt into another group of text prompts: ``a table", ``a table with a red table cloth", and ``a table with yellow tulips". We synthesize the base subject ``table" first and then bind each attachment to it. Our approach relieves the concept bleeding problem of multiple attachments. We further demonstrate the necessity of base subjects with ablation analysis in Appendix \ref{ablation}.

\begin{algorithm}[t]
\caption{Isolated Diffusion for Multiple Attachments.}
\label{algo:algo1}
\begin{algorithmic}[1]

\REQUIRE ~~\\
\textbf{Fixed Models:} CLIP text encoder $\text{CLIP}_{\text{text}}$, image decoder $\mathcal{D}$, pre-trained diffusion model $\epsilon_{\theta}$.\\
\textbf{Input:} Prompt list $\mathcal{P}=[\text{p}_{ucon},\text{p}_{base},\text{p}_1,...,\text{p}_k]$, random Gaussian noise $x_T$, scale hyperparameter $\lambda$.\\
\textbf{Output:} Generated image $X$.\\

\STATE $[\text{c}_{ucon},\text{c}_{base},\text{c}_1,\ldots,\text{c}_k] \leftarrow \text{CLIP}_{\text{text}}(\mathcal{P})$. \\

\FOR{$t=T,T-1,\ldots,1$}
\STATE$\epsilon_{ucon}\leftarrow\epsilon_{\theta}(x_t,t,\text{c}_{ucon})$, \hspace{1pt} $\epsilon_{base}\leftarrow\epsilon_{\theta}(x_t,t,\text{c}_{base})$
\STATE $\hat{\epsilon}(x_t,t) \leftarrow \epsilon_{ucon}+\lambda (\epsilon_{base}-\epsilon_{ucon})$
\FOR{$i=1,\ldots,k$}
\STATE $\epsilon_i \leftarrow \epsilon_{\theta}(x_t,t,\text{c}_k)$
\STATE $\hat{\epsilon}(x_t,t) += \lambda(\epsilon_i-\epsilon_{base})$
\ENDFOR
\STATE Denoise $x_t$ with $\hat{\epsilon}(x_t,t)$ to $x_{t-1}$.
\ENDFOR
\STATE Feed $x_0$ to decoder $\mathcal{D}$ to generate $X$.
\end{algorithmic}
\end{algorithm}

\subsection{Isolated Diffusion for Multiple Subjects}
\label{33}
Apart from multiple attachments, another common case of concept bleeding occurs when generating multiple subjects. Taking a simple text prompt composed of two subjects like ``A dog next to a cat" as an example, SD still encounters concept bleeding as shown by the sample of ``A cat next to a cat" in the middle part of Fig. \ref{pipeline}. Different subjects in an image can interfere with each other, leading to text-image inconsistency.

\begin{algorithm}[t]
\caption{Isolated Diffusion for Multiple Subjects}
\label{algo:algo2}
\begin{algorithmic}[1]
\REQUIRE ~~\\
\textbf{Fixed Models:} CLIP text encoder $\text{CLIP}_{\text{text}}$, image decoder $\mathcal{D}$, pre-trained diffusion model $\epsilon_{\theta}$, $\text{YOLO}$ and $\text{SAM}$ models.\\
\textbf{Input:} Prompt list $\mathcal{P}=[\text{p}_{ucon},\text{p}_{con},\text{p}_1,...,\text{p}_k]$, random Gaussian noise $x_T,x_{\epsilon}$, scale hyperparameter $\lambda$. \\
\textbf{Output:} Generated image $X$.\\

\STATE $[\text{c}_{ucon},\text{c}_{con},\text{c}_1,\ldots,\text{c}_k] \leftarrow \text{CLIP}_{\text{text}}(\mathcal{P})$

\STATE Generate $X_0$ with $\epsilon_{\theta}$ based on $\text{c}_{ucon}$ and $\text{c}_{con}$.

\STATE Object detection on $X_0$ with YOLO
\IF{Detection results match text prompt with enough confidence:}

\STATE $X \leftarrow X_0$
\ELSE
\STATE Segment subjects in $X_0$ with $\text{SAM}$ using the center points of bounding boxes as prompts and get masks $[\mathcal{M}_1,\ldots,\mathcal{M}_k]$, then assign masks to each subject.
\FOR{$t=T,T-1,\ldots,T_{layout}$}
\STATE $\epsilon_{ucon} \leftarrow \epsilon_{\theta}(x_t,t,\text{c}_{ucon})$, $\epsilon_{con} \leftarrow \epsilon_{\theta}(x_t,t,\text{c}_{con})$
\STATE $\epsilon = \epsilon_{ucon} + \lambda * (\epsilon_{con}-\epsilon_{ucon})$
\STATE Denoise $x_t$ with $\epsilon$ to $x_{t-1}$.
\ENDFOR
\FOR{$i=1,\ldots,k$}
\STATE $\mathcal{M} \leftarrow \mathop{\cup}^{n\neq i}_{1\leq n \leq k} \mathcal{M}_n$
\STATE $x^i_{T_{lay}}=x_{T_{lay}}\otimes(1-\mathcal{M})+ x_{\epsilon}\otimes\mathcal{M}$
\ENDFOR
\FOR{$t=T_{lay},\ldots,1$}
\STATE $\epsilon_{ucon}^0 \leftarrow \epsilon_{\theta}(x_t,t,\text{c}_{ucon})$,\hspace{1pt} $\epsilon_{con}^0 \leftarrow \epsilon_{\theta}(x_t,t,\text{c}_{con})$
\STATE $\epsilon^0 = \epsilon_{ucon}^0 + \lambda(\epsilon_{con}^0-\epsilon_{ucon}^0)$
\STATE $\hat{\epsilon}(x_t,t) \leftarrow \epsilon^0\otimes(1-\mathop{\cup}_{1\leq n \leq k} \mathcal{M}_{n})$

\FOR{$i=1,\ldots,k$}
\STATE $\epsilon_{ucon}^i \leftarrow \epsilon_{\theta}(x^i_t,t,\text{c}_{ucon})$, $\epsilon_{con}^i \leftarrow \epsilon_{\theta}(x^i_t,t,\text{c}_{i})$
\STATE $\epsilon^i = \epsilon^i_{ucon} + \lambda(\epsilon^i_{con}-\epsilon^i_{ucon})$
\STATE Denoise $x^i_t$ with $\epsilon^i$ to $x^{i}_{t-1}$
\STATE $\hat{\epsilon}(x_t,t) += \epsilon^i \otimes \mathcal{M}_i$
\ENDFOR
\STATE Denoise $x_t$ with $\hat{\epsilon}(x_t,t)$ to $x_{t-1}$
\ENDFOR
\STATE Feed $x_0$ to decoder $\mathcal{D}$ to generate $X$.
\ENDIF

\end{algorithmic}
\end{algorithm}

Similar to Isolated Diffusion for multiple attachments, we also separate the denoising processes of different subjects to overcome the concept bleeding problem of multiple subjects. We propose to revise the concept bleeding samples produced by SD models. Compared with multi-attachment synthesis, we need to maintain the layouts of multiple subjects, which is difficult to solve automatically.

We depend on pre-trained detection and segmentation models to identify the positions of subjects by producing masks for them. Here we employ YOLO and SAM as examples to illustrate our approach. Firstly, we use YOLO to detect subjects in synthesized samples and determine whether concept bleeding occurs (e.g., detection inconsistent with subjects in text prompts or detection with low confidence). If so, we synthesize masks for each subject with SAM using the center points of bounding boxes as axes prompts. Then we assign the masks with split text prompts of each subject. The method of subject-mask assignment is flexible. Taking the right part of Fig. \ref{examples0} as an example, we use various subject-mask assignments and revise the synthesized sample in different ways.

In the revision process, we first denoise the initial latents for $T_{lay}$ steps to synthesize the layouts of all the subjects, which is determined in the early denoising steps \cite{bar2023multidiffusion,couairon2022diffedit}. Then we replace the regions in $x_{T_{lay}}$ corresponding to other subjects with random noises to avoid mutual interference between different subjects for following denoising steps under split conditions. Taking the sample in Fig. \ref{pipeline} as an example, we denote $\mathcal{M}_{dog}$ and $\mathcal{M}_{cat}$ as the masks of the dog and cat to be synthesized.  In the denoising process of each subject, we replace the regions of other subjects with random noises. From the view of the attention mechanism, we manipulate the query maps by replacing the regions of other subjects with random noises to obtain attention maps for each subject individually. The latents to be denoised under the conditions of ``A dog" and ``A cat" can be defined as follows:
\begin{align}
    x_{T_{lay}}^{1} &= x_{T_{lay}} \otimes (1-\mathcal{M}_{cat}) + \epsilon \otimes \mathcal{M}_{cat} \\
    x_{T_{lay}}^{2} &= x_{T_{lay}} \otimes (1-\mathcal{M}_{dog}) + \epsilon \otimes \mathcal{M}_{dog}
\end{align}
where $\epsilon$ and $\otimes$ represent random noises and the element-wise multiplication of tensors. Then they are denoised separately with the predicted noises as follows:
\begin{align}
    \epsilon^{i}=(1-\lambda)\epsilon_{\theta}(x_t^i,t,c_{ucon})+\lambda\epsilon_{\theta}(x_t^i,t,c_{i}).
\end{align}
At every time step $t$, we denoise $x_{t}$ using the combination of predicted noises in different regions, including foreground and background, segmented by the masks as follows:
\begin{align}
    \epsilon^0 &= ((1-\lambda)\epsilon_{\theta}(x_t,t,c_{ucon})+\lambda\epsilon_{\theta}(x_t,t,c_{con})) \\
    \epsilon^i &= ((1-\lambda)\epsilon_{\theta}(x_t^i,t,c_{ucon})+\lambda\epsilon_{\theta}(x_t^i,t,c_{i})) \\
    \hat{\epsilon}(x_t,t) &= \epsilon^0 \otimes(1-\cup_{1\leq n\leq k}\mathcal{M}_n) + \sum_{i=1}^{k}\epsilon^i \otimes \mathcal{M}_i.
\end{align}
where $k$ represents the number of subjects in the synthesized sample (e.g., $k=2$ for ``A dog next to a cat").

The pseudo-code of Isolated Diffusion for multiple subjects is provided in Algorithm \ref{algo:algo2}. Compared with the SD inference process, we split text prompts into a group of simpler text prompts for each subject and introduce the open-source YOLO and SAM models to fix the concept bleeding problem for multiple subjects. There are alternative choices to erase the attention on other subjects with random noises. Ablations of noise adding strategies are provided in Appendix \ref{ablation}.

\begin{figure*}[t]
    \centering
    \includegraphics[width=1.0\linewidth]{ 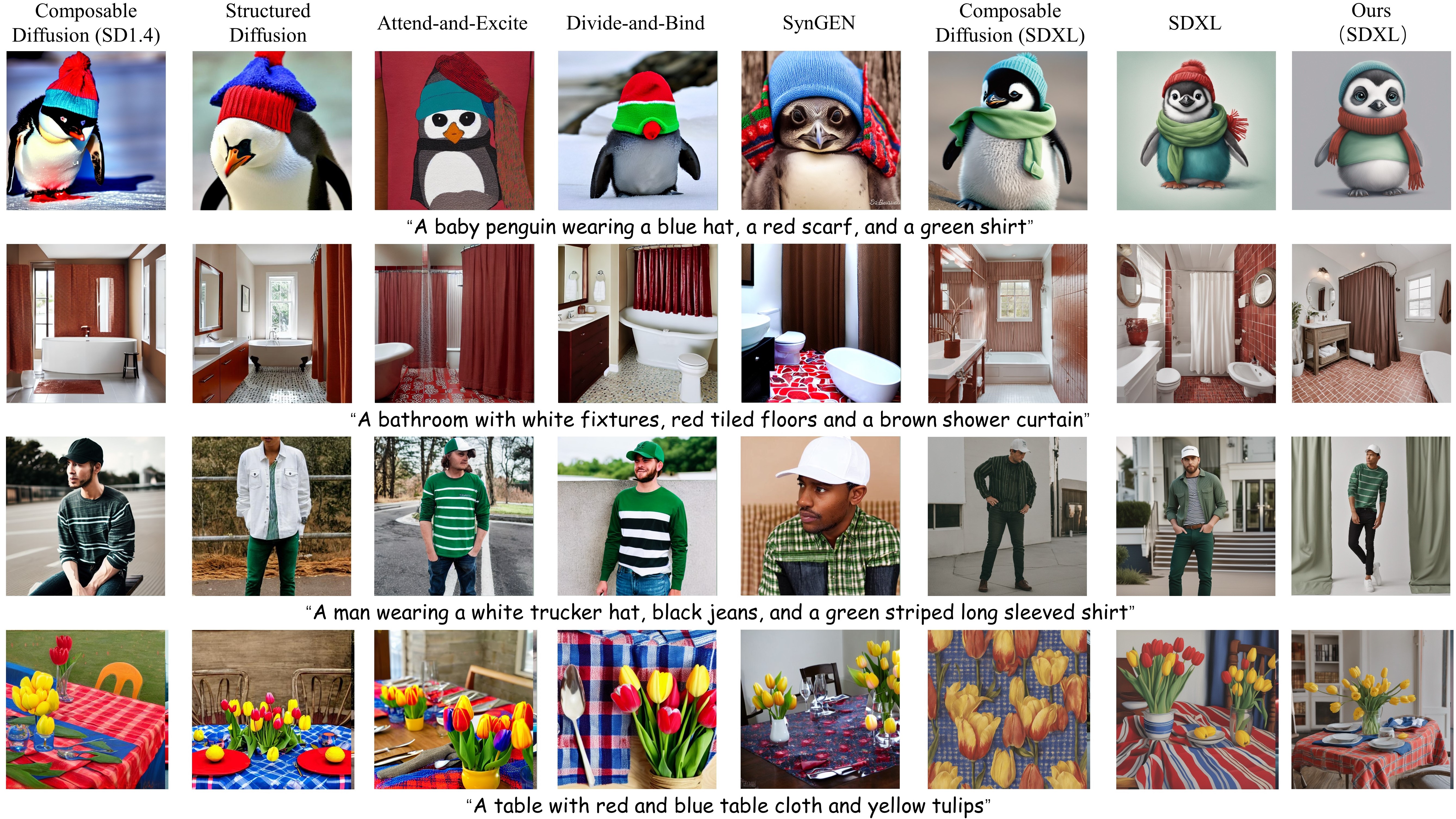}
    \caption{Qualitative comparison between our approach and baselines using text prompts of multiple attachments. Isolated Diffusion achieves the best text-image consistency among all the methods and maintains high fidelity similar to SDXL. }
    \label{result1}
\end{figure*}

\section{Experiments}
\label{sec:experiments}
\textbf{Basic Setups} The proposed Isolated Diffusion is mainly implemented with SDXL \cite{podell2023sdxl}. Our approach is training-free and the inference processes are conducted on a single NVIDIA RTX A6000 GPU. The max diffusion step $T$ is 1000. We sample images with the DPM-solver \cite{lu2022dpm} scheduler using 50 spaced time steps. The refiner introduced in SDXL is employed to denoise samples in the last 10\% time steps. For multi-subject generation, we empirically set the time step $T_{lay}$ as 700-800 to maintain the original layouts of subjects. We follow SDXL to set the scaling of condition $\lambda$ as 5 for all the experiments. The YOLOv8x \cite{wang2023yolov7} and SAM ViT-H \cite{kirillov2023segment} models are employed in multi-subject generation.

\textbf{Datasets} We compile some text prompts from MS-COCO \cite{lin2014microsoft} and synthesize additional text prompts following specific formats of multiple concepts using GPT4 \cite{schick2023toolformer} as supplements. The formats for multiple concepts can be approximately concluded as follows: 1) a [subject] with [attachment1], [attachment2], and [attachment3], 2) a [subject A] and a [subject B], and 3) a [adjective A] [subject A] and a [adjective B] [subject B], 4) a [subject A] with [attachment A1], [attachment A2] and a [subject B] with [attachment B1], [attachment B2].

\textbf{Baselines} We employ Composable Diffusion \cite{liu2022compositional}, Structured Diffusion \cite{feng2022training}, Attend-and-Excite \cite{chefer2023attend}, Divide-and-Bind \cite{li2023divide}, and SynGEN \cite{rassin2024linguistic} as baselines. SD1.4 and SDXL \cite{podell2023sdxl} as used as foundation models. All baselines are designed based on SD1.x or SD2.x models. However, SDXL uses different distributions of transformer blocks. Therefore, we implement most baselines with SD1.4 and provide the results of our approach based on both SD1.4 and SDXL. Conform \cite{meral2023conform} tackles similar tasks but is not open-source yet. Therefore, we do not include it for comparison.

% We employ Structured Diffusion \cite{feng2022training} and Attend-and-Excite \cite{chefer2023attend}, which are designed based on SD1.4 as baselines. Composable Diffusion \cite{liu2022compositional} based on SD1.4 and SDXL are used as baselines as well. In addition, samples produced by SDXL are provided as reference.

\begin{figure*}[h]
    \centering
    \includegraphics[width=1.0\linewidth]{ 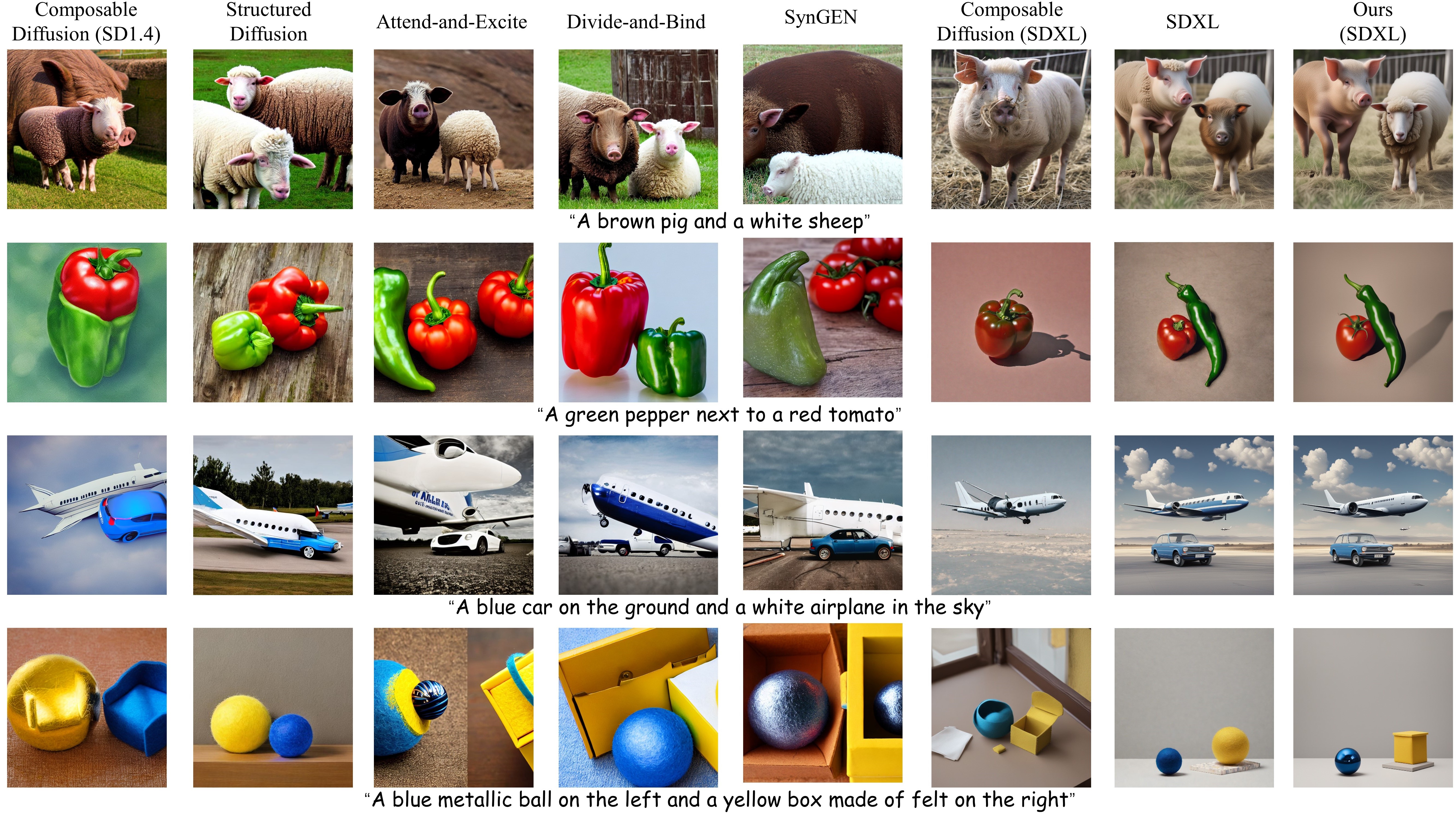}
    \caption{Qualitative comparison between our approach and baselines using text prompts of multiple subjects. Isolated Diffusion revises the samples of SDXL to fix the problem of concept bleeding and achieves outstanding text-image consistency. }
    \label{result2}
\end{figure*}

 \begin{figure*}[h]
    \centering
    \includegraphics[width=1.0\linewidth]{ 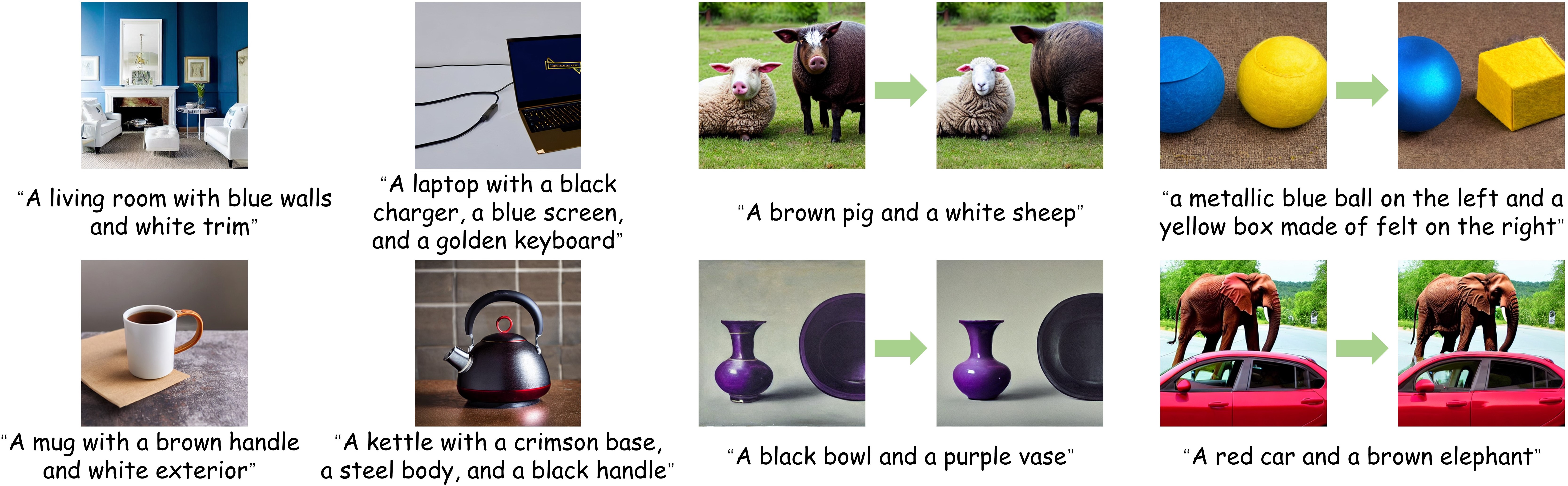}
    \caption{Additional visualized samples of Isolated Diffusion based on SD1.4 for multiple attachments (the left two columns) and multiple subjects (the right two columns). In the two columns on the right, we provide the concept bleeding samples of SD1.4 on the left and the revised results of our approach on the right.}
    \label{result_add3}
\end{figure*}

\begin{figure*}[h]
    \centering
    \includegraphics[width=1.0\linewidth]{ 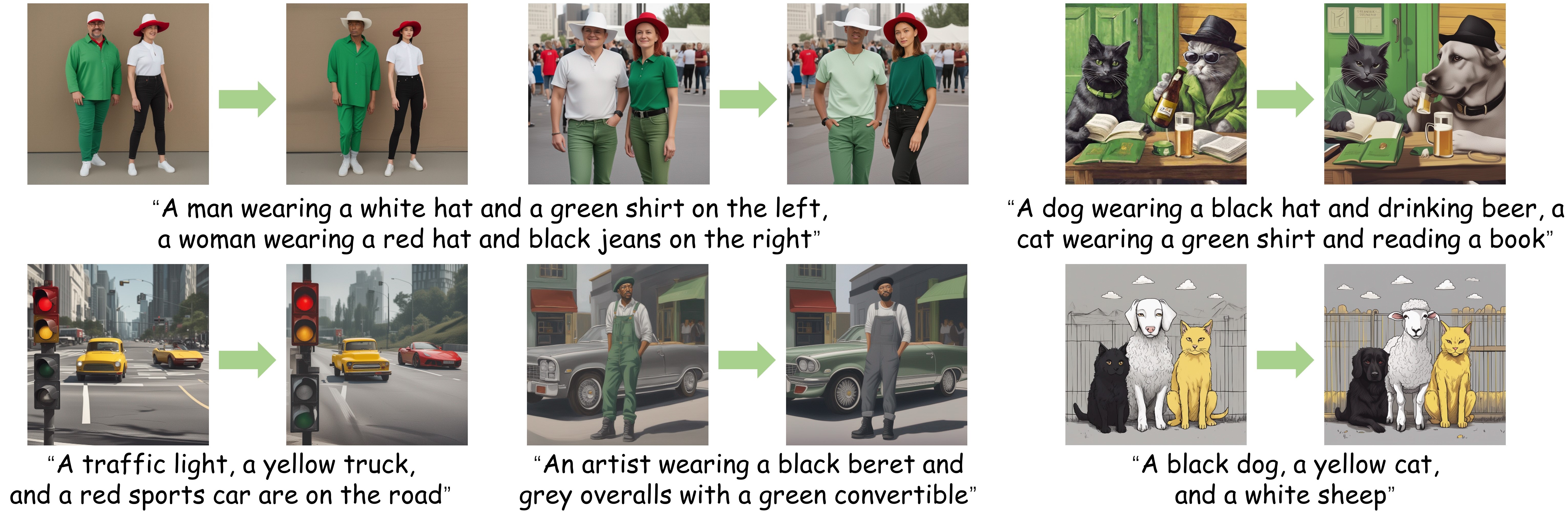}
    \caption{Qualitative comparison between samples produced by SDXL and revised by our approach in more complex scenes, e.g., more kinds of subjects or two subjects with various attachments bound to each one.}
    \label{result3}
\end{figure*}

\subsection{Qualitative Evaluation}
 We provide visualized samples accompanied by text prompts to evaluate the text-image matching degree qualitatively. We provide results of multi-attachment synthesis in Fig. \ref{result1}. Our approach produces the best text-image consistent results with high fidelity. In detail, our approach avoids the concept bleeding of color and material descriptions bound to different attachments. Taking the first row as an example, our approach produces a baby penguin and synthesizes each attachment accurately. Baselines either neglect the ``green shirt" or merge ``green" to ``hat" or ``scarf".

\begin{table*}[t]
   \centering
   \begin{tabular}{l|l|c|c|c|c|c|c}
   \hline \multirow{2}{*}{Foundation Model} & \multirow{2}{*}{Method} & \multicolumn{2}{c|}{CLIP-TEXT ($\uparrow$)} & \multicolumn{2}{c|}{BLIP-VQA ($\uparrow$)} & \multicolumn{2}{c}{mGPT-CoT ($\uparrow$)} \\ \cline{3-8}
   & & Attachments & Subjects & Attachments & Subjects & Attachments & Subjects \\
    \hline
    \multirow{6}{*}{SD1.4} & Composable Diffusion \cite{liu2022compositional} & $0.310$ & $0.310$  & 0.356 & 0.382 & 0.812 & 0.688\\
    & Structured Diffusion \cite{feng2022training} & $0.316$ & $0.310 $ & $0.379$ & $0.496$ & 0.815 & 0.765\\
    &Attend-and-Excite \cite{chefer2023attend} & $0.315 $ & $0.323$ & $0.430$ & $0.589$ & 0.815 &  0.795\\
    &Divide-and-Bind \cite{li2023divide} & $0.322$ & $0.331$ & $0.451$ & $0.549$ & 0.825 & 0.771\\
    &SynGEN \cite{rassin2024linguistic} & $0.320$ & $0.338$ & $0.515$ & $0.594$ & 0.845 & $0.805$  \\
    &Ours & $\pmb{0.328}$ & $\pmb{0.346}$ & $\pmb{0.532}$ & $\pmb{0.627}$ & $\pmb{0.854}$ & $\pmb{0.842}$\\
    \hline
    \multirow{3}{*}{SDXL} & Composable Diffusion \cite{liu2022compositional} & $0.318  $ & $0.321$ & $0.443$ & $0.396$ & 0.825 & 0.765\\
    &SDXL \cite{podell2023sdxl}& $0.319 $ & $0.329$ & $0.507$ & $0.542$ &  0.844 & 0.766\\
    &Ours & $\pmb{0.332 }$ & $\pmb{0.354 }$ & $\pmb{0.571}$ & $\pmb{0.699}$ & $\pmb{0.879}$ & $\pmb{0.861}$\\
    \hline
    \end{tabular}
    \caption{Quantitative evaluation of text-image consistency for baselines and our approach. Results of different approaches averaged over 50 text prompts of multiple attachments or multiple subjects. Our approach outperforms baselines on all the benchmarks. We split the results according to the foundation model choice for fair comparison.}
    \label{quantitative}
\end{table*}

%We compare our approach with baselines using a series of text prompts.
% according to the text prompt

In Fig. \ref{result2}, we show visualized samples of multiple subjects. Unlike the strategy for multiple attachments, Isolated Diffusion for multiple subjects is designed to fix the concept bleeding problem in samples produced by SD \cite{rombach2022high} models. Therefore, we provide samples of SDXL and samples revised by our approach. We observe that baselines tend to generate samples of merging concepts. Given ``a brown pig and a white sheep" as the text prompt, Structured Diffusion \cite{feng2022training} synthesizes two sheep with different colors; Attend-and-Excite \cite{chefer2023attend} gets pigs with sheep ears; Divide-and-Bind \cite{li2023divide} and SynGEN \cite{rassin2024linguistic} get sheep with a pig nose; SDXL \cite{podell2023sdxl} produces a sheep having a brown head and pig nose; Composable Diffusion \cite{podell2023sdxl} fuses pig and sheep directly. Our approach avoids the mutual interference between different subjects and synthesizes a brown pig and a white sheep appropriately. Moreover, baselines suffer from subjects merging or missing and wrongly assigned attachments between various subjects. Our approach draws support from YOLO and SAM models to separate the synthesis of each subject. It achieves the best text-image consistency and maintains image layouts and high generation quality. Visualized samples of our approach based on SD1.4 are shown in Fig. \ref{result_add3}. More samples are added in Appendix \ref{appendix_result}.

We follow prior works to focus mainly on two-subject scenes. We also explore samples of more complex scenes, such as three subjects or two subjects with multiple attachments bound to each one in Fig. \ref{result3}. We combine Isolated Diffusion for multiple attachments and multiple subjects to revise the mutual interference between various concepts and achieve better text-image consistency in such complex scenes.

\subsection{Quantitative Evaluation}
\label{42}
We use 50 text prompts for multi-attachment and another 50 text prompts for multi-subject evaluation. We first encode the generated samples and text prompts to embeddings with CLIP \cite{radford2021learning} image and text encoders. Then we compute the cosine similarity between text and image embeddings as the CLIP-text metric. Besides, we employ the other two benchmarks designed for text-image consistency evaluation in multi-concept generation: BLIP-VQA and mGPT-CoT \cite{huang2024t2i}, which employs BLIP \cite{li2022blip} and MiniGPT4 \cite{zhu2023minigpt} to understand synthesized samples. The results are split into two groups based on different foundation models SD1.4 and SDXL. As shown in Table \ref{quantitative}, our approach achieves state-of-the-art results on all the benchmarks, demonstrating its strong capability of maintaining text-image consistency for multi-concept synthesis.

\subsection{User Study}
\label{43}

We employ 75 participants in the user study and provide 125 questions for everyone. The ages of users range from 18 to 55. Users are asked to consider both text-image consistency and generation quality given text prompts as reference.

\begin{table}[t]
\centering
\small
\begin{tabular}{l|c|c}
\hline Method & Attachments & Subjects\\
\hline
Composable Diffusion (SD1.4) \cite{liu2022compositional} & 2.18\% & 2.27\% \\
Structured Diffusion (SD1.4) \cite{feng2022training}& 7.07\% & 5.78\% \\
Attend-and-Excite (SD1.4) \cite{chefer2023attend}& 1.42\% & 11.51\% \\
Divide-and-Bind (SD1.4) \cite{li2023divide}& 2.67\% & 12.44\% \\
SynGEN (SD1.4) \cite{rassin2024linguistic} & 18.71\% & 17.73\%\\
Composable Diffusion (SDXL) \cite{liu2022compositional} & 8.00\% & 4.93\% \\
SDXL \cite{podell2023sdxl}& 13.87\% & 5.07\% \\
Ours (SDXL) & \textbf{46.09\%} & \textbf{40.27\%} \\
\hline
\end{tabular}
\caption{User study results. We achieve the highest support rates in two subsets of multiple attachments and subjects.}
\label{HUMAN1}
\end{table}

In the first 50 questions, users are provided with 8 options of our approach and baselines, including Structured Diffusion \cite{feng2022training}, Attend-and-Excite \cite{chefer2023attend}, Divide-and-Bind \cite{li2023divide}, SynGEN \cite{rassin2024linguistic}, SDXL \cite{podell2023sdxl}, and Composable Diffusion \cite{liu2022compositional} based on SD1.4 and SDXL and asked to choose the best sample. We synthesize samples based on SD1.4 and SDXL with fixed seeds separately for fair comparison. We choose 25 text prompts for each task (multiple attachments and multiple subjects). For multi-subject generation, we provide revised samples using axes prompts for the subjects provided by users. We count up the votes for every approach and report the results of the user study by percentage in Table \ref{HUMAN1}. For multiple attachments, our approach outperforms baselines and takes almost half of the votes. SynGEN achieves higher approval ratings than other baselines. Attend-and-Excite and Divide-and-Bind are not designed for multi-attachment synthesis. For multiple subjects, our approach achieves the highest approval rate of 40.27\%. Attend-and-Excite and Divide-and-Bind get apparently better results in multi-subject synthesis. Baselines like Structured Diffusion and Composable Diffusion fail to produce compelling results for multiple subjects. Some baselines can synthesize samples of high fidelity for some relatively simple prompts, making it hard for our approach to obtain overwhelming advantages among all the approaches.

In the next 50 questions, we implement methods including Structured Diffusion, Attend-and-Excite, Divide-and-Bind, SynGEN, and our approach with SD1.4 and ask users to choose the best one. The results are reported in Table \ref{HUMAN2}. Our approach still achieves clear advantages over baselines in both multi-attachment and multi-subject synthesis.

In the last 25 questions, users are asked to identify whether the revised samples of multiple subjects produced by our approach are better than the original ones or if the revised samples are similar to the original ones. Our approach gains the support rate of 82.93\% while 3.64\% votes think that the original samples are better, and other votes remain neutral.

% \begin{table}[ht]
% \centering
% \setlength\tabcolsep{1.2mm}{
% \small
% \begin{tabular}{l|c}
% \hline Method & Object Detection Score ($\uparrow$)\\
% \hline
% Composable Diffusion (SD1.4) & $0.724 \pm 0.092$ \\
% Structured Diffusion (SD1.4) & $0.876 \pm 0.071$  \\
% Attend-and-Excite (SD1.4) &  $1.152 \pm 0.044$  \\
% Ours (SD1.4) & $\pmb{1.404 \pm 0.047}$ \\ \hline
% Composable Diffusion (SDXL) &  $0.701 \pm 0.045 $\\
% SDXL &  $0.945 \pm 0.050$  \\
% Ours (SDXL) &  $\pmb{1.579 \pm 0.036}$  \\
% \hline
% \end{tabular}}
% \caption{Object detection scores of different approaches averaged over 50 text prompts of multiple subjects. Standard deviations are computed across 500 samples of 50 prompts.}
% \label{yolo2}
% \end{table}

\begin{table}[t]
\centering
\small
\begin{tabular}{l|c|c}
\hline Method & Attachments & Subjects\\
\hline
Structured Diffusion (SD1.4) \cite{feng2022training} & 13.87\% & 4.22\% \\
Attend-and-Excite (SD1.4) \cite{chefer2023attend} & 4.18\% & 14.27 \% \\
Divide-and-Bind (SD1.4) \cite{li2023divide} & 4.49\% & 17.02 \% \\
SynGEN (SD1.4) \cite{rassin2024linguistic} & 25.20\% & 25.33 \% \\
Ours (SD1.4) & \textbf{52.71\%} & \textbf{39.16\%} \\
\hline
\end{tabular}
\caption{User study implemented with SD1.4.}
\label{HUMAN2}
\end{table}

\subsection{Ablation Analysis}
\label{ablation}
\begin{figure}[t]
    \centering
    \includegraphics[width=1.0\linewidth]{ 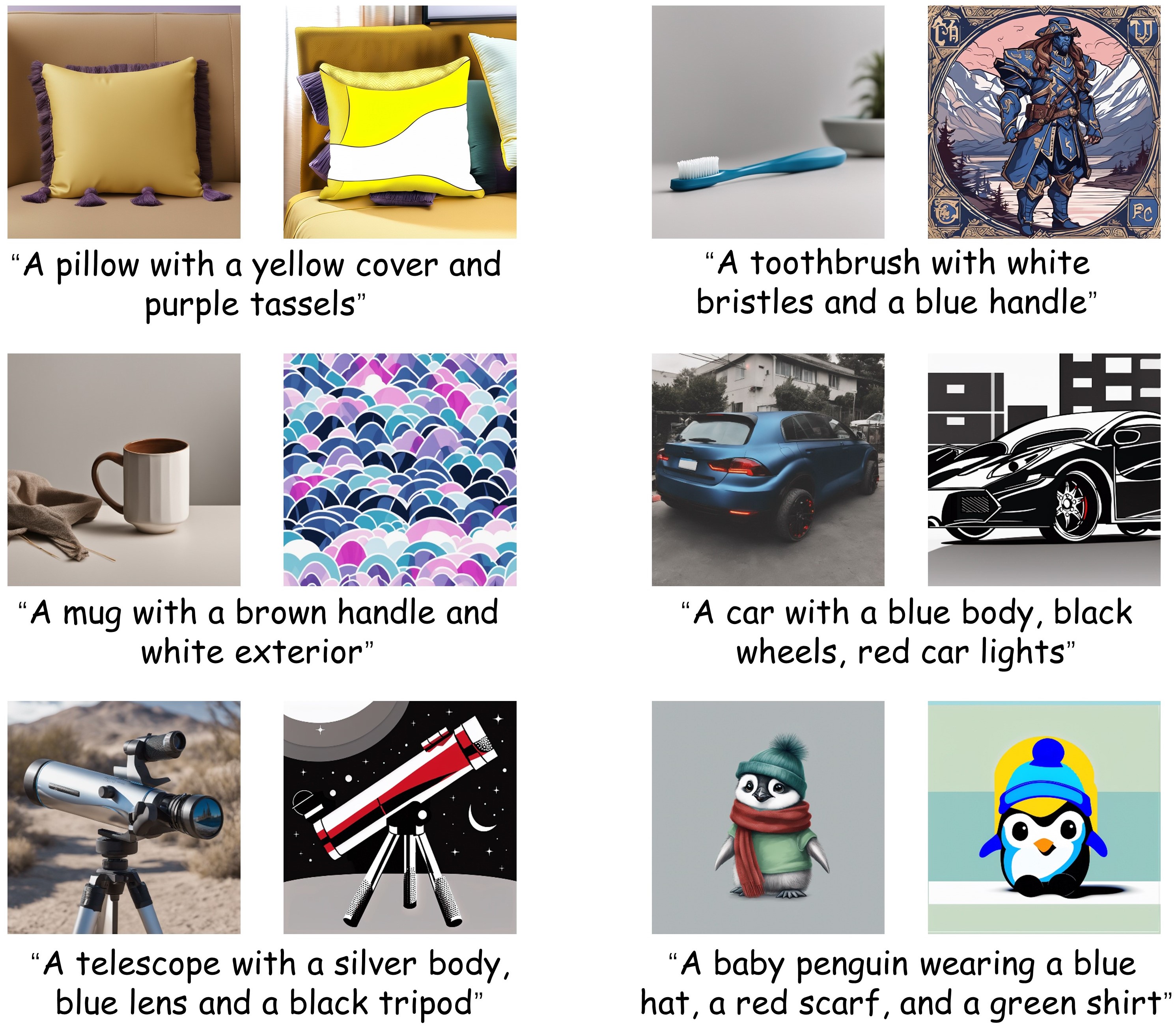}
        \caption{Qualitative ablations of Isolated Diffusion for multiple attachments. Samples of Isolated Diffusion are shown on the left for every text prompt. Samples of Isolated Diffusion w/o using $\text{p}_{base}$ are shown on the right. Input noises are fixed.}
    \label{ablation0}
\end{figure}

We first ablate Isolated Diffusion for multiple attachments with different methods to combine the attachments with the subject. As illustrated in Sec. \ref{32}, we use the variance between noises predicted with the prompt of the base subject $\text{p}_{base}$ and the prompt of each attachment bound to the base subject $\text{p}_{k}$. Naturally, there exists another method to isolate the denoising processes of multiple attachments. We can abandon using $\text{p}_{base}$ and add up noises predicted with $\text{p}_{k}$ directly. In this way, Eq. \ref{eq2} degenerates to Eq. \ref{eq22} as follows:

\begin{equation}
    \begin{aligned}
    \hat{\epsilon}(x_t,t) & =  \epsilon_{\theta}(x_t,t,c_{ucon})   \\
    &+ \sum_{i=1}^k \lambda(\epsilon_{\theta}(x_t,t,c_{i})-\epsilon_{\theta}(x_t,t,c_{ucon})) \label{eq22}
\end{aligned}
\end{equation}
We provide ablations of our approach and this method with visualized samples synthesized from fixed noise inputs in Fig. \ref{ablation0}.  It can be seen that without $\text{p}_{base}$, it becomes hard to obtain realistic samples. Most samples are abstract and inconsistent with text prompts. Some totally meaningless samples can be found as well. It further validates the effectiveness of our approach and the necessity of generating samples of multiple attachments based on the prompt of the base subject $\text{p}_{base}$.

\begin{figure*}[t]
    \centering
    \includegraphics[width=1.0\linewidth]{ 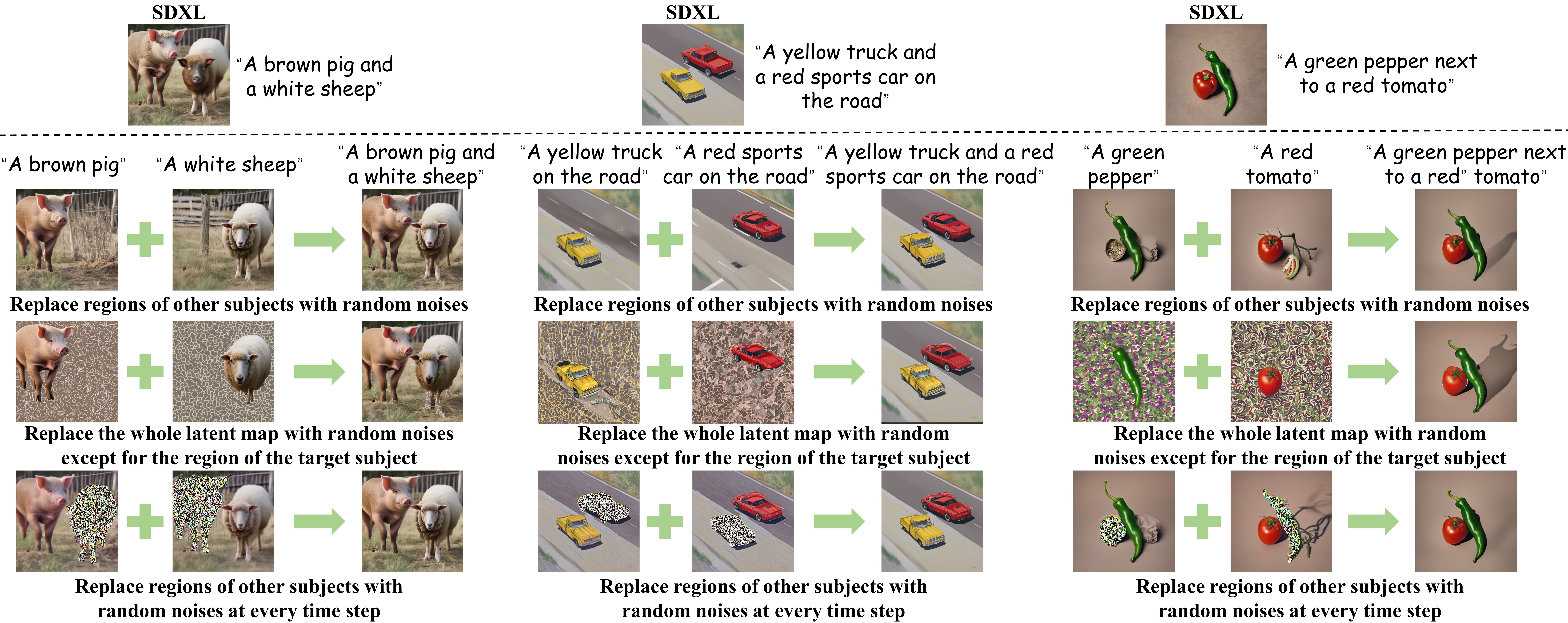}
    \caption{Qualitative ablations for different noise adding strategies in multi-subject synthesis.}
    \label{ablation1}
\end{figure*}

% \begin{figure}[t]
%     \centering
%     \includegraphics[width=1.0\linewidth]{ ablation1.jpg}
%     \caption{Qualitative ablations using the text prompt ``A brown pig and a white sheep" as an example.}
%     \label{ablation1}
% \end{figure}

% \begin{figure}[t]
%     \centering
%     \includegraphics[width=1.0\linewidth]{ ablation3.jpg}
%     \caption{Qualitative ablations using the text prompt ``A yellow truck and a red sports car on the road" as an example.}
%     \label{ablation3}
% \end{figure}

%  \begin{figure}[t]
%     \centering
%     \includegraphics[width=1.0\linewidth]{ ablation2.jpg}
%     \caption{Qualitative ablations using the text prompt ``A green pepper next to a red tomato" as an example.}
%     \label{ablation4}
% \end{figure}

Then we ablate Isolated Diffusion for multiple subjects with different methods to erase the attention on the other subjects. In Sec. \ref{33}, we replace the regions of other subjects with random noises at the time step $T_{lay}$ (method A) and denoise each subject individually in the following steps. Here we provide the other two methods: 1) replace the whole latent with random noises except for the region of the target subject (method B), 2) replace regions of other subjects with random noises at every time step after $T_{lay}$ (method C).

We employ several text prompts and provide qualitative results in Fig. \ref{ablation1}. We provide denoised samples of each subject and the final outputs, all of which are refined by the refiner of SDXL in the last 10\% timesteps. As shown in the visualized samples, all three methods achieve better text-image consistency with isolated denoising processes of various subjects. However, it can be seen that it is difficult for method B to fuse subjects into the background naturally, leading to degraded generation quality. Method C achieves results similar to method A. However, it can be seen that keeping the regions of other subjects as random noises influences the generation quality of target subjects in certain cases. It leads to some unreasonable details, like the red sporting car in Fig. \ref{ablation1}. Therefore, we choose to use method A in Isolated Diffusion.

%To sum up, our approach achieves obvious superiority over all the other baselines on multi-concept generation in the user study.

\subsection{Comparison with MultiDiffusion}
\begin{figure}[t]
    \centering
    \includegraphics[width=1.0\linewidth]{ 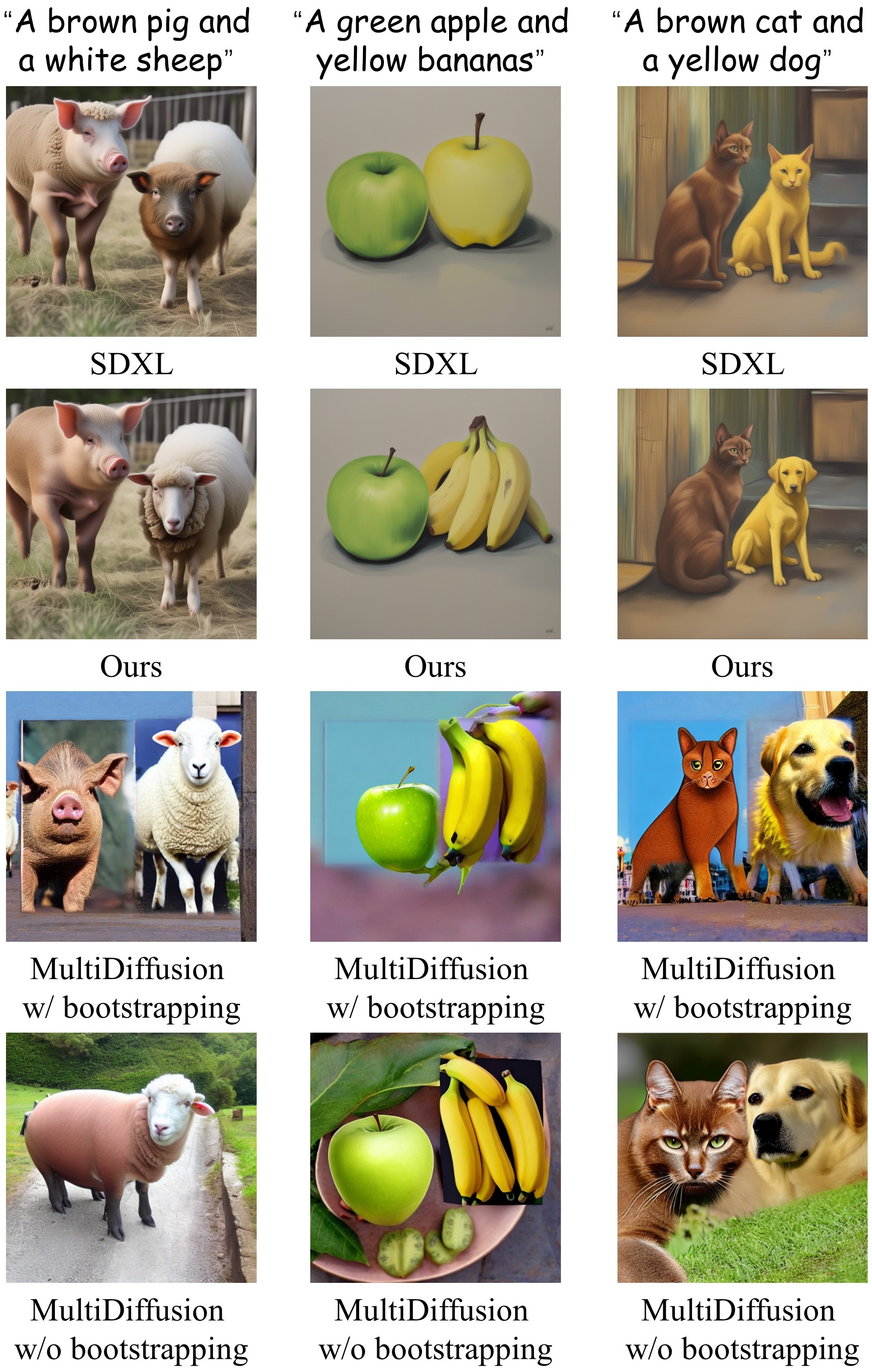}
    \caption{Qualitative comparison between our approach and MultiDiffusion. We place coarse masks for two subjects in the left and right parts of images as conditions for MultiDiffusion.}
    \label{multi}
\end{figure}

MultiDiffusion \cite{bar2023multidiffusion} is a controllable generation method compatible with region-based image generation given masks for foreground subjects. It also splits text prompts for multi-subject generation. Here we add detailed comparison between MultiDiffusion and our approach. Our key idea is to erase attention on other subjects when denoising each subject individually with split text prompts. When denoising each subject, we replace the regions of other subjects in latents with random noises. MultiDiffusion denoises the same latents with split text prompts and fuses regions segmented by masks with closed-form formulas to get the full image. It doesn't take actions to erase attention on other subjects. We provide several samples produced by MultiDiffusion in Fig. \ref{multi} using coarse masks placed in the left and right parts of images and provide results of our approach for comparison. MultiDiffusion suffers two typical problems in multi-subject synthesis. It may merge two subjects into one or produce subjects in disharmonious styles. In contrast, our approach achieves more reasonable results.

%with isolated guidance applied to revise subjects synthesized with stable diffusion models

\section{Limitations}
\label{limitations}
Despite the compelling results achieved by our approach, it still has some limitations. Our approach is designed to solve the concept bleeding problem of current SD models. However, it fails when SD models neglect subjects or YOLO fails to detect enough number of subjects. For example, given a text prompt of ``A cat and a dog", our approach cannot deal with samples of only one cat or dog. Fortunately, SDXL has made great progress in avoiding the concept missing problem. For example, we use SDXL to synthesize hundreds of samples with text prompts like ``A cat and a frog" and find almost all the samples get results without subject missing. As for YOLO, synthesized samples generally have higher resolution and are clearer than natural images, making foreground subjects easier to be detected in most cases. We can also adjust confidence thresholds to obtain appropriate results. In addition, it is hard for YOLO to deal with unseen subjects. Alternative detectors or human feedback can be applied as replacements to deal with such cases.

\section{Conclusion}
\label{sec:conclusion}
This work introduces Isolated Diffusion to handle the well-known ``concept bleeding" problem of modern text-to-image SD models. Our approach is designed to deal with the mutual interference between different attachments and subjects in multi-concept generation. Different from recent works optimizing cross-attention maps or latents, we explore a novel route to isolate different concepts more intuitively. In detail, we isolate the denoising processes of various attachments and subjects using split text prompts as conditions. For multiple attachments, we synthesize each attachment bound to the same subject separately. For multiple subjects, we depend on pre-trained detection and segmentation models to identify image layouts and generate each subject separately. Our approach is training-free and compatible with any SD models. We conduct extensive experiments and demonstrate the effectiveness of our approach with clear advantages over existing methods in the user study. Our work takes a further step towards better text-image consistency in multi-concept synthesis. %Moreover, our approach is not constrained by specific network architectures and can be compatible with more powerful diffusion models in the future.

\bibliographystyle{IEEEtran}
\bibliography{tvcg}

\vfill

\clearpage

\appendix
\begin{table*}[t]
    \centering
    \begin{tabular}{c|c}
        \hline
        Text Prompt & Prompt List $\mathcal{P}$ \\
        \hline
      \makecell[c]{A bathroom with white fixtures, red tiled floors \\ and a brown shower curtain} &  \makecell[c]{\textbf{A bathroom}, A bathroom with a white fixtures, \\ A bathroom with red tiled floors, \\ A bathroom with a brown shower curtain}  \\
      \hline
        \makecell[c]{A car with black wheels, a blue body, and red car lights} & \makecell[c]{\textbf{A car}, A car with black wheels, \\ A car with a blue body, A car with red car lights} \\
     \hline
        \makecell[c]{A toothbrush with white bristles and a blue handle}  & \makecell[c]{\textbf{A toothbrush}, A toothbrush with white bristles, \\ A toothbrush with a blue handle} \\
    \hline
        \makecell[c]{A table with red and blue table cloth and yellow tulips} & \makecell[c]{\textbf{A table}, A table with red and blue table cloth, \\
        A table with yellow tulips} \\
     \hline
        \makecell[c]{A yellow truck and  a red sports car on the road} & \makecell[c]{A yellow truck on the road, A red sports car on the road } \\
     \hline
        \makecell[c]{A metallic blue ball on the left and \\ a yellow box made of felt on the right} & \makecell[c]{A metallic blue ball on the left, \\A yellow box made of felt on the right} \\
    \hline
        \makecell[c]{A blue car on the ground and a white airplane in the sky} & \makecell[c]{A blue car on the ground,  A white airplane in the sky} \\
    \hline
        \makecell[c]{A green apple and  yellow bananas} & \makecell[c]{A green apple,  Yellow bananas} \\
     \hline
        \makecell[c]{A man wearing a white hat and a green shirt on the left, \\ a woman wearing a red hat and black jeans on the right} & \makecell[c]{\textbf{A man on the left}, \\ A man wearing a white hat on the left,\\ A man wearing a green shirt on the left, \\ \textbf{A woman on the right}, \\ A woman wearing a red hat on the right, \\ A woman wearing black jeans on the right} \\
     \hline
    \end{tabular}
    \caption{Prompt lists (separated by commas, the right part) used as inputs for Isolated Diffusion corresponding to the original text prompts (the left part). The text prompts of the base subjects $\text{p}_{base}$ in multi-attachment generation are highlighted in \textbf{bold text}. Only the decomposed text prompts are shown here. The unconditional prompt $\text{p}_{ucon}$, and the original text prompt $\text{p}_{con}$ are omitted. GPT4 \cite{schick2023toolformer} is employed to automate the text split process.}
    \label{prompt_list}
\end{table*}

\subsection{Comparison with Image Editing Methods}
\label{appendix_compare_related}
As illustrated in Sec. \ref{sec:related}, image editing methods based on text-to-image diffusion models \cite{bau2021paint,nichol2022glide,avrahami2023blended,mao2023guided,brooks2023instructpix2pix,couairon2022diffedit,avrahami2022blended} tackle tasks different from this work. They aim to manipulate parts of given images, while our approach aims to fix the concept bleeding problem in multi-concept generation. We use SAM \cite{kirillov2023segment} to get masks of subjects in Isolated Diffusion for multiple subjects. Some image editing methods like Blended Diffusion \cite{avrahami2023blended,avrahami2022blended} and DiffEdit \cite{couairon2022diffedit} conduct image editing based on masks as well. Here we add detailed comparison.

\begin{figure*}[t]
    \centering
    \includegraphics[width=1.0\linewidth]{ 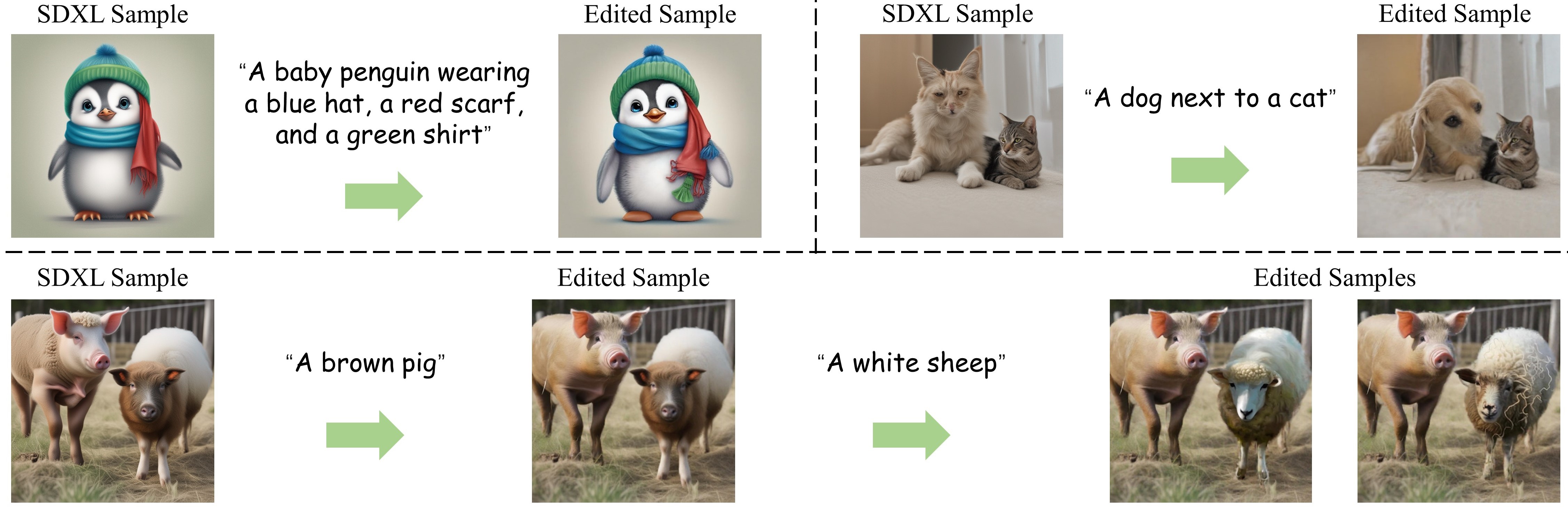}
    \caption{Image editing examples produced by Blended Latent Diffusion \cite{avrahami2023blended} based on SDXL \cite{podell2023sdxl}. We use Blended Latent Diffusion to edit samples produced by SDXL. Blended Latent Diffusion can optimize the concept bleeding problem of multi-subject synthesis but still struggles to achieve harmonious results. It is not qualified for multi-attachment synthesis.}
    \label{result_blended}
\end{figure*}

Blended Diffusion marks the regions to be edited with user-given masks and denoise the whole image with text prompts of the edited content. In every denoising step, Blended Diffusion replaces the unedited regions with original images or compressed latents added with noises. Similarly, DiffEdit synthesizes masks for the subjects to be edited and denoises the whole image with text prompts of the edited content. DiffEdit shares the same method of maintaining unedited regions with Blended Diffusion. These methods may fail to produce samples that look realistic as a whole since they do not provide accurate guidance for the regions to be edited with given text prompts. Instead, they denoise the whole image with text prompts of edited parts directly and replace the unedited regions with the original image. Besides, they also struggle to deal with multi-attachment scenes like SD models. We provide samples produced by Blended Latent Diffusion in Fig. \ref{result_blended}. It fails to deal with multi-attachment scenes and struggles to synthesize high-quality samples for multiple subjects. DiffEdit is not open-source yet, but some failed cases of multi-concept scenes are provided in its supplementary.

Isolated Diffusion is designed to achieve text-image consistency in multi-concept generation. It isolates the generation process of each subject by replacing the regions of other subjects with random noises and guides the diffusion model to focus on a single subject with its corresponding text prompt. It composes the whole latent with regions of each subject and denoises the background with complete text prompts.

%Their performance relies on the choice of attention maps. Therefore, it becomes hard to extend them to SDXL, which has a lot more cross-attention maps in its huge UNet network. Therefore, we implement them based on SD1.4 with the officially released code.  Composable Diffusion and our approach are not bound with certain network architectures.
\subsection{More Details of Implementation}
We provide several examples of prompt lists $\mathcal{P}$ corresponding to multi-concept text prompts in Table \ref{prompt_list}. For multiple attachments, we obtain the prompt of the subject and bind each attachment to this subject separately. For multiple subjects, we split the text of each subject directly. As shown in the last row of Table \ref{prompt_list}, we can combine these two methods to deal with more complex prompts containing multiple attachments bound to multiple subjects. We employ GPT4 \cite{schick2023toolformer} to automate this text split process.

 In Isolated Diffusion for multiple subjects, we empirically find that time step $T_{lay}$ between 700 and 800 is appropriate to keep the layouts of subjects and achieve effective revision of the subjects to avoid concept bleeding. We recommend larger $T_{lay}$ to reserve more denoising steps for greater changes. For example, when Isolated Diffusion aims to change subjects into other subjects with different shapes (e.g., sphere $\rightarrow$ box), more steps are needed to achieve such significant changes.

 Besides, it is worth noting that YOLO \cite{wang2023yolov7} models are not always qualified for detecting target subjects, especially when the target subjects are not included in its training datasets. In such cases, we recommend users to provide axes prompts with alternative detection models or human feedback for the SAM \cite{kirillov2023segment} model to segment the synthesized subjects. In addition, our approach is promising to benefit from the SAM model using text prompts only, which has yet to be open-source.

 \subsection{Additional Visualized Samples}
\label{appendix_result}
\begin{figure*}[h]
    \centering
    \includegraphics[width=1.0\linewidth]{ 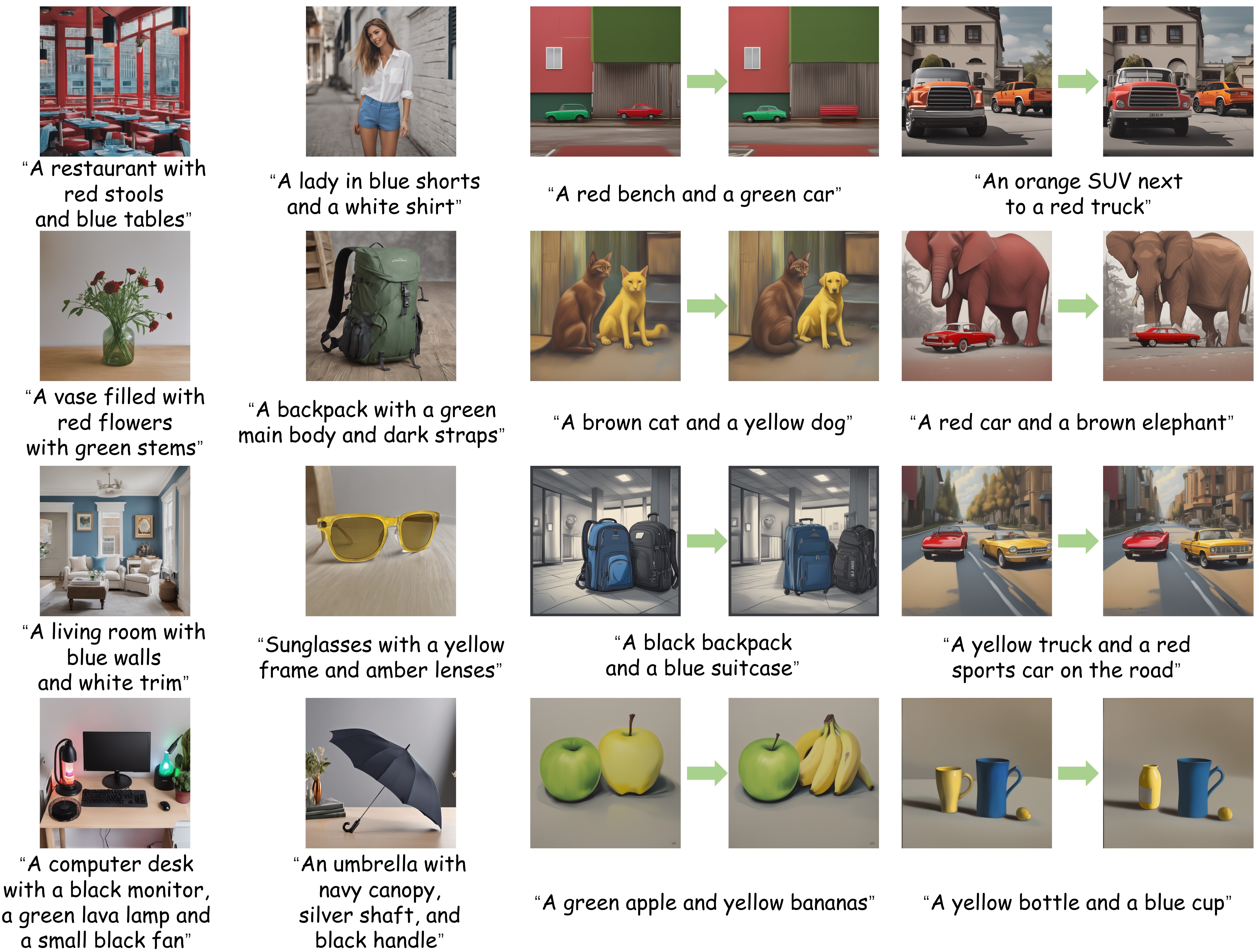}
    \caption{Additional visualized samples of Isolated Diffusion for multiple attachments (the left two columns) and multiple subjects (the right two columns). In the right two columns, we provide the concept bleeding samples of SDXL \cite{podell2023sdxl} on the left and the revised results of our approach on the right.}
    \label{result_add}
\end{figure*}

We provide additional multi-attachment and multi-subject samples produced by our approach in Fig. \ref{result_add} as supplements to Sec. \ref{sec:experiments}.

\end{document}